\pdfoutput=1

\documentclass[11pt]{article}
\usepackage{amsmath}
\usepackage[preprint]{acl}
\usepackage{graphicx}
\usepackage{enumitem}
\usepackage{times}
\usepackage{latexsym}
\usepackage{booktabs} 
\usepackage[T1]{fontenc}
\usepackage{tabularx}
\newcolumntype{Y}{>{\centering\arraybackslash}X}
\usepackage{pifont}
\usepackage[utf8]{inputenc}
\usepackage{amssymb}
\usepackage{microtype}
 \usepackage{multirow} 
\usepackage{inconsolata}

\title{Enhancing EEG-to-Text Decoding through Transferable Representations from Pre-trained Contrastive EEG-Text Masked Autoencoder}

\author{Jiaqi Wang\textsuperscript{1,2},
        Zhenxi Song\textsuperscript{1}\thanks{Corresponding author},
        Zhengyu Ma\textsuperscript{2},
        Xipeng Qiu\textsuperscript{3},
        Min Zhang\textsuperscript{1},
        Zhiguo Zhang\textsuperscript{1,2}\thanks{Corresponding author} \\[1ex] 
        \textsuperscript{1}School of Computer Science and Technology, Harbin Institute of Technology Shenzhen, China \\
        \textsuperscript{2} Peng Cheng Laboratory, China \\
        \textsuperscript{3}School of Computer Science, Fudan University, China \\
        \texttt{\{songzhenxi, zhiguozhang\}@hit.edu.cn}\\
}

\begin{document}
\maketitle

\begin{abstract}
    Reconstructing natural language from non-invasive electroencephalography (EEG) holds great promise as a language decoding technology for brain-computer interfaces (BCIs). 
    However, EEG-based language decoding is still in its nascent stages, facing several technical issues such as: 
    1) Absence of a hybrid strategy that can effectively integrate cross-modality (between EEG and text) self-learning with intra-modality self-reconstruction of EEG features or textual sequences;
    2) Under-utilization of large language models (LLMs) to enhance EEG-based language decoding.
    To address above issues, we propose the \textbf{C}ontrastive \textbf{E}EG-\textbf{T}ext \textbf{M}asked \textbf{A}uto\textbf{e}ncoder (\textbf{CET-MAE}), a novel model that orchestrates compound self-supervised learning across and within EEG and text through a dedicated multi-stream encoder.
    Furthermore, we develop a framework called \textbf{E2T-PTR} (\textbf{E}EG-\textbf{to}-\textbf{T}ext decoding using \textbf{P}retrained \textbf{T}ransferable \textbf{R}epresentations), which leverages pre-trained modules alongside the EEG stream from CET-MAE and further enables an LLM (specifically BART) to decode text from EEG sequences. 
    Comprehensive experiments conducted on the popular text-evoked EEG database, ZuCo, demonstrate the superiority of E2T-PTR, which outperforms the baseline framework in ROUGE-1 F1 and BLEU-4 scores by 8.34\% and 32.21\%, respectively. Our proposed pre-trained EEG-Text model shows the potential to improve downstream tasks involving EEG and text. This opens up promising avenues for its application in inner speech BCI paradigms, meriting further investigation.
    
\end{abstract}

\section{Introduction}

Decoding natural language from non-invasive brain recordings with electroencephalography (EEG) is an emerging topic that holds promising benefits for patients suffering from cognitive impairments or language disorders. Thanks to the burgeoning development of pre-trained large language models (LLMs)~\cite{zhao2023survey}, the potential of using an open vocabulary to decode human brain activity has been gradually unlocked. Specifically, through the commendable text understanding and generation capabilities of cutting-edge LLMs~\cite{touvron2023llama,ouyang2022training}, translating complex spatio-temporal EEG signals into nuanced textual representations, which is known as EEG-to-Text, is being achieved. Compared to conventional paradigms of brain-computer interfaces (BCIs), such as motor imagery (MI)~\cite{al-saeghDeepLearningMotor2021}, steady-state visual evoked potential (SSVEP)~\cite{wang2016benchmark}, and P300~\cite{cecotti2010convolutional}, EEG-to-Text can convey much more intended commands from the human brain to computers, and thus presents a more extensive range of applications. 
Its potential as a novel and powerful BCI paradigm suggests it could contribute to advancements in the field of imagined or inner speech BCIs. \par
\par

\begin{figure}[!t]
    \centering
    \includegraphics[width=0.5\textwidth]{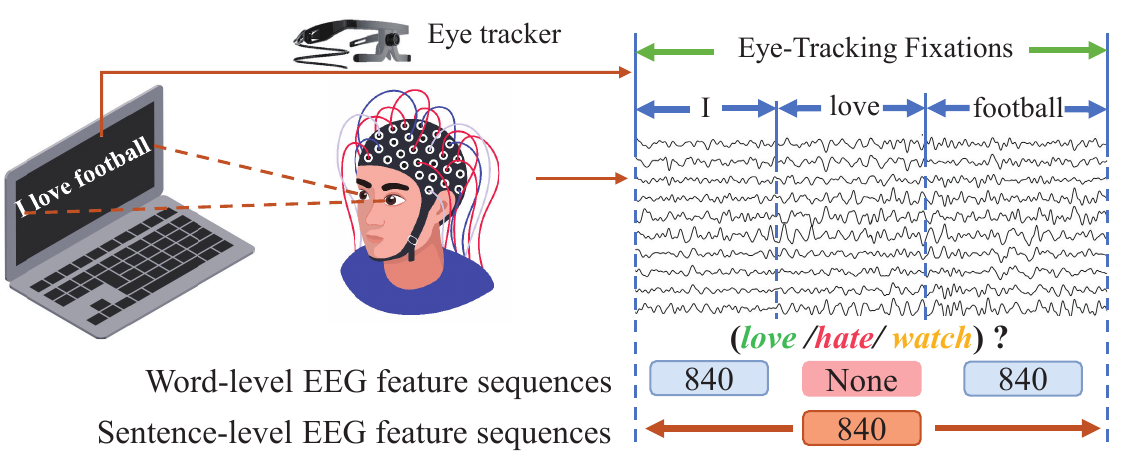}
       \caption{\textbf{Text-evoked EEG Recording in ZuCo datasets.} Participants' EEG and eye-tracking data are simultaneously recorded during natural reading to capture text-evoked brain activity.}
    \label{Figure1}
\end{figure}


Several existing EEG studies~\cite{li2022multi, yi2024learning} were focused on developing specialized pre-trained models for EEG only, aiming to extract universal semantic representations from the human brain. However, the pre-trained model bridging EEG and text has been ignored, which may be important to enhance the representation learning for inter-modality conversion~\cite{bai2023dreamdiffusion}.
This motivates us to develop a hybrid model to orchestrate compound pre-trained representations across and within EEG and text.
This endeavor faces the core challenge:
\textit{How to bridge the semantic gap between EEG and text while establishing an implicit mapping in the latent representation space?} 
Responding to this challenge, we focus on self-supervised learning (SSL), because of its great capability in multi-modal representation learning~\cite{chenContextAutoencoderSelfsupervised2024}.
Contrastive learning is one of the important SSL strategies, learning semantic-level representations across modalities (as CLIP does for language and image) ~\cite{radford2021learning}.
Masked modeling methods exhibit significant capability of intra-modality self-reconstruction, such as BERT~\cite{devlin2019bert} in nature language processing and masked autoencoder (MAE)~\cite{he2022masked} in computer vision. \par
Inspired by the above prevailing SSL strategies, we propose a novel pre-trained model to align EEG and text, Contrastive EEG-Text Masked Autoencoder (CET-MAE), as shown in Figure\ref{Figure2}(a). 
CET-MAE integrates contrastive learning and masked signal modeling through a dedicated multi-stream encoder. It effectively learns pre-trained representations of EEG and text by balancing the latent embeddings represented by self-reconstruction and the semantic-level aligned embeddings of text tokens and text-evoked EEG features.
In terms of masked signal modeling, CET-MAE implements a high mask ratio (specifically, 75\%) on both EEG and text data, presenting a meaningful challenge for the model to handle an increased amount of missing information during the reconstruction phase. This setting not only enhances the model's understanding of individual modality but also facilitates cross-modal interactions and support.\par

Furthermore, to make the most of LLMs' capability in language understanding and generation as well as to fully use pre-trained representations learned by CET-MAE, we introduce a new EEG-to-Text decoding framework, EEG-to-Text using Pre-trained Transferable Representations (E2T-PTR). E2T-PTR utilizes pre-trained modules alongside the EEG stream from CET-MAE and further adopts the BART~\cite{lewis2020bart} to decode language from EEG sequences. By transferring the pre-trained representations from CET-MAE, E2T-PTR significantly enhances EEG-to-Text decoding, surpassing both the baseline and state-of-the-art (SOTA) methods.\par


Our main contributions are summarised below:
\begin{itemize}
    \item Introducing CET-MAE, the first pre-trained EEG-text model for EEG-based language decoding. CET-MAE integrates the reconstruction of text and EEG features with semantic alignment, forming a multi-stream SSL framework for both intra-modality and cross-modality representation learning.
    \item Developing a new EEG-to-Text framework via E2T-PTR. The new E2T-PTR framework can leverage CET-MAE's pre-trained EEG representations and the capabilities of LLMs (BART) for text generation.
    \item Conducting extensive EEG-to-Text experiments on three, four, and five reading tasks in ZuCo. 
    Results demonstrate that our framework outperforms previous works, offering valuable insights into leveraging pre-trained transferable representations to enhance EEG-to-text decoding.

\end{itemize}

\section{Related Works}

\subsection{Self-supervised Representations Learning}
Multimodal self-supervised representation learning aims to explore the interactions between different modalities to produce semantically generalizable representations for downstream tasks.  \par
In recent years, there have been substantial progresses across various modalities, such as vision-language pre-training~\cite{zhao2023mamo,linSmaugSparseMasked2023}. A range of existing methods rely on contrastive learning, which can effectively draw closer to the global representations of matched pairs in latent spaces with semantic-level self-supervised constraints. But contrastive learning sometimes tends to overlook the self-information of individual modalities, particularly at more granular levels. 
On the other hand, multimodal masked signal modeling integrates cross-modality self-learning with intra-modality self-reconstruction, focusing on reconstructing one modality from another. This approach may help the model learn the associations between modalities. However, it may lead to an excessive emphasis on fine-grained details, potentially weakening the overall cross-modality correlation and causing issues such as insensitivity to whether the inputs are matched pairs. A series of recent works, such as CMAE~\cite{huangContrastiveMaskedAutoencoders2023}, CAV-MAE~\cite{gong2022contrastive} and SimVTP~\cite{ma2022simvtp}, have already successfully integrated both contrastive learning and masked signal modeling so that their complement advantages can be utilized. \par
Our work draws inspiration from the above SSL methods but with a novel strategy. In the proposed CET-MAE, the utilization of both text and EEG streams not only achieves an explicit contrastive learning objective to capture global coordination but also avoids erroneous learning processes. Meanwhile, the utilization of the joint stream can facilitate the information interaction between modal-specific embeddings to achieve masked signal modeling effectively. To the best of our knowledge, this is the first EEG-to-Text masked autoencoder that attempts to establish transferable representation learning between EEG and text.

\begin{figure*}[!ht]
    \centering
    \includegraphics[width=0.98\textwidth]{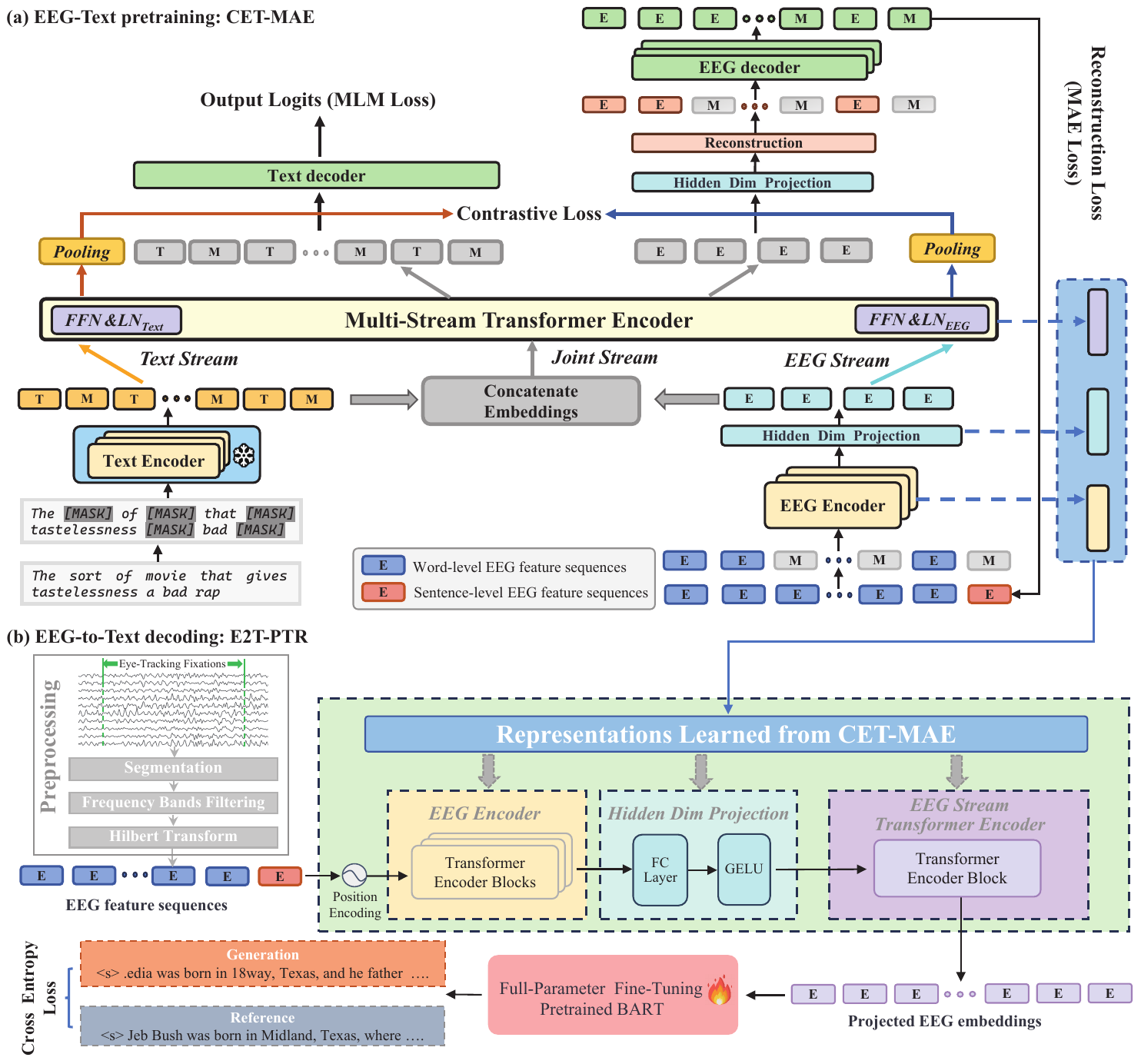} %
    \caption{
    Illustration of the proposed EEG-text pre-training model (CET-MAE) and EEG-to-Text decoding framework (E2T-PTR).
    (a) \textbf{CET-MAE Model:}
    CET-MAE features modality-specific autoencoders with a masking strategy for text and EEG features, complemented by a multi-stream transformer encoder that orchestrates self-reconstruction and cross-modality semantic alignment, enhancing representation learning for EEG semantic decoding.
    (b) \textbf{E2T-PTR Framework:}
    E2T-PTR transfers both word- and sentence-level EEG representations extracted from CET-MAE's pre-trained modules, further facilitating text generation through the BART.
    }
    \label{Figure2}
\end{figure*}

\subsection{Open Vocabulary EEG-to-Text Decoding}
Previous works~\cite{nieto2022thinking,kamble2023spectral} on EEG-to-Text have been severely confined by a limited number of (several or tens of) words in terms of vocabulary size. These closed-vocabulary efforts primarily focused on recognizing low-level linguistic features, such as individual words or syllables. However, these works can hardly capture more complex, high-level semantic and contextual aspects of language.\par
The development of LLMs has significantly enhanced the field of EEG-based text decoding. The first work using LLM~\cite{wang2022open} integrates an additional EEG encoder to align the pre-trained BART for EEG-to-Text, providing important inspiration for subsequent works. 
C-SCL~\cite{feng2023aligning} employs curriculum learning to effectively mitigate the discrepancy between subject-dependent and semantic-dependent EEG representations in EEG-to-Text translation. DeWave~\cite{duan2024dewave} uses a quantized variational encoder to convert continuous EEG signals into discrete sequences, alleviating the reliance on eye fixations.
Despite advancements, prior efforts struggled to bridge the complex semantic gap between EEG and text on an open-vocabulary scale. Our proposed CET-MAE aims to tackle this challenge. Additionally, our E2T-PTR framework transfers CET-MAE's representations and leverages the BART to achieve superior text generation outcomes.



\section{Methods}
\label{sec:methods}

\subsection{Preliminary}
\label{sec:intro}
\textbf{ZuCo benchmark dataset}. For our work, we use the ZuCo1.0~\cite{hollenstein2018zuco} and ZuCo2.0~\cite{hollensteinZuCoBenchmarkCrosssubject2023} datasets, which contain the EEG and eye tracking data during five natural reading tasks. The corpus for sentiment reading (SR) task v1.0 comes from the movie reviews. The corpus for the remaining four tasks is sourced from Wikipedia and comprises two versions each of Natural Reading (NR) and Task-Specific Reading (TSR), specifically NR v1.0, NR v2.0, TSR v1.0, and TSR v2.0. The word-level EEG was recorded and aligned by the eye-tracking fixations, and the sentence-level EEG was recorded during the entire reading procedure. We follow the preprocessing and dataset splits established by baseline work~\cite{wang2022open}.\par

\textbf{Natural masking ratios of EEG feature sequences}. Our investigation reveals the word-level contextual EEG presentations in ZuCo datasets are severely corrupted due to missing eye-tracking fixations, leading to mismatches between EEG raw data and text, as shown in Figure\ref{Figure1}. This misalignment leads to fragmented word-level EEG feature sequences, which fails to capture the cohesive semantics of entire sentences and inevitably complicates the representations learning of EEG and text.\par
Different from previous works, we concatenate the word-level EEG features and the sentence-level EEG features as our EEG feature sequences \textbf{\textit{E}} as 
\begin{equation}
\textbf{E} = [E_{word1},E_{word2},..,E_{wordN},E_{sentence}].
\end{equation}

Incorporating sentence-level EEG features offers several benefits.
First, it provides a holistic view of EEG sequences, enriching the interpretation of overall sentence semantics.
Secondly, it acts as a form of data augmentation, which can mitigate the issue of data incompleteness, thereby alleviating semantic discrepancies caused by the misalignment between word-level EEG and text. To provide a clearer overview, we have presented the detailed statistics of the natural masking ratio (NMR) of EEG feature sequences under three categories of reading task combinations in Appendix ~\ref{appendixa}.\par

\textbf{Definitions in EEG-to-Text Decoding}. Given a sequence of EEG features \textbf{\textit{E}} as the input to the model \textbf{\textit{M}}, the aim is to decode the ground-truth word tokens \textbf{\textit{W}} from open-vocabulary \textbf{\textit{V}} via \textbf{\textit{M}}. These corresponding EEG-Text pairs {$\left \langle \textit{\textbf{E}}, \textit{\textbf{W}} \right \rangle $} are collected during natural readings. \par
During the testing phase, the model \textbf{\textit{M}} operates with an implicit understanding of the ground-truth word tokens \textbf{\textit{W}}. Its primary objective remains to decode the EEG feature sequences \textbf{\textit{E}} to generate an output that closely matches tokens \textbf{\textit{W}}. This involves the model generating the sequence of words with the highest probability within the probability distribution \textbf{\textit{P}} of the \textbf{\textit{V}}.\par

\subsection{EEG-Text Masking}
\label{sec:intro1}
We perform random masking on the text tokens, followed by processing with BERT. For EEG masking, we adopted the following settings. Word-level EEG feature sequences are randomly masked, while sentence-level EEG feature sequences are compulsorily masked. This aims to force the model to fully reconstruct the contextual semantics within the sentence-level EEG feature sequences. \par


\subsection{CET-MAE Encoder}
As illustrated in Figure~\ref{Figure2}(a), the CET-MAE model needs to extract the embeddings of text and EEG separately and then feed the embeddings into the multi-stream transformer encoder to learn the cross-modal representations.\par

\textbf{Text encoder}.  We utilize the pre-trained encoder-decoder model BART as the text encoder. Due to the suitable capabilities in natural language understanding and generation~\cite{liContrastGenerationMake2022}, we opt to freeze weights of the BART \footnote{\url{https://huggingface.co/facebook/bart-large}} encoder to maintain its high-level language comprehension from the last hidden states.  Firstly, the text tokens are converted into high-quality text embeddings with positional encoding by BART. The learnable embeddings are then used to replace the masked word tokens.\par 

\textbf{EEG encoder}. The EEG encoder is designed as a Multi-layer Transformer Encoder )~\cite{vaswani2017attention} to capture the temporal relationships from EEG sequences with spatial and frequency features in each token.
A learnable linear projection layer is employed to transform the EEG embeddings from the EEG encoder, aligning their dimensions with those of the text embeddings.\par

\textbf{Multi-stream Transformer encoder}. The pivotal design of this module lies in the integration of EEG, text, and the joint streams.
We implement the dual-modality streams for EEG-text contrastive learning, especially using a specialized head for each modality.
It is equipped with the layer normalization (LN) and the feed-forward network (FFN) enabling the production of embeddings that preserve their unique properties\cite{gong2022contrastive}. Notably, we control the learning process to ensure that learnable vectors at masked positions do not enter into the text stream, thereby preventing the inclusion of misleading contrastive feedback. Equally crucial for the two reconstruction tasks, the joint stream is utilized to facilitate the integration of the embeddings from both text and EEG modalities. This design aims to deepen the interaction and enhance the cooperation between EEG and text, fostering a more effective learning synergy.\par

\subsection{CET-MAE Decoder}
We apply a lightweight Transformer encoder as the EEG decoder. For EEG reconstruction tasks, EEG embeddings are first mapped to the original dimensions through a learnable linear projection layer. Subsequently, EEG embeddings with learnable masked tokens are inserted back into their original positions. The final EEG embeddings added to the positional embeddings are fed into the EEG decoder. Since the text encoder has already encoded the masked tokens and captured their positional information within the text, we employ a learnable linear projection layer as the text decoder to predict the masked text tokens.\par


\subsection{CET-MAE Training Objectives}
CET-MAE is pre-trained by three objectives: 
(1) Masked Text Modeling  ({$L_{T}$}): it aims to predict the masked text tokens by utilizing hybrid representations that integrate information from both textual and EEG embeddings. 
(2) Masked EEG Modeling ({$L_{E}$}): it learns to reconstruct the original EEG feature sequences, especially predicting masked word- and sentence-level features based on hybrid representations, where the error is measured by mean square error (MSE).
(3) EEG-Text Contrastive Learning ({$L_{CL}$}): it involves a process where the corresponding EEG and text representations are computed by separate global average pooling layers. The objective is to bring the aligned pairs (matched EEG and text embeddings) closer together while pushing unpaired ones further apart. Our goal {$L$} is minimizing is the summation of these three learning objectives:
\begin{equation}
L = \lambda _{T} \cdot L_{T} + \lambda _{E} \cdot L_{E} + \lambda _{CL} \cdot L_{CL}
\end{equation}

\subsection{E2T-PTR Framework}
The proposed E2T-PTR is illustrated in Figure~\ref{Figure2}(b). It can be summarized into the following key points.\par
\textbf{Word-sentence level input tokens.} We add the sentence-level EEG features as our input tokens. As detailed in \ref{sec:intro}, concatenating the sentence-level EEG feature sequences as the last token can effectively alleviate the incoherent contextual semantics due to gaps in word-level EEG features.

\textbf{Effective transfer capability.} We investigate how to effectively transfer the cross-modality representations learned from the CET-MAE to downstream tasks such as EEG-to-Text decoding. The E2T-PTR employs a synergy of the following critical components: the EEG encoder, the linear projection layer, and the EEG-stream transformer encoder, all of which are integral components as outlined within the CET-MAE. For the LLM backbone, we also apply the BART which excels at natural language generation tasks.



\textbf{Fine-tuning strategy.} We fine-tune all parameters of E2T-PTR during the training phase. The weights of CET-MAE are first loaded into the EEG encoder, the linear projection layer, and the EEG-stream transformer encoder. As the linguistic backbone of E2T-PTR, the BART is also fully fine-tuned to improve its ability to generate fine-grained text tokens from EEG embeddings. \par

\begin{table*}[!ht]
\centering
\resizebox{\textwidth}{!}{%
\begin{tabular}{l|c|llll|lll}
\toprule
\textbf{Method} & \multicolumn{1}{c|}{\textbf{Training}} & \multicolumn{4}{c|}{\textbf{BLEU-N(\%)}} & \multicolumn{3}{c}{\textbf{ROUGE-1(\%)}} \\
 & \multicolumn{1}{c|}{\textbf{Sample}} & \textbf{N=1} & \textbf{N=2} & \textbf{N=3} & \textbf{N=4} & \textbf{P} & \textbf{R} & \textbf{F} \\
\midrule
EEG2Text~\cite{wang2022open} & 10710 & 40.1 & 23.1 & 12.5 & 6.8 & 31.7 & 28.8 & 30.1 \\
DeWave~\cite{duan2024dewave} & 10710 & 41.35 & 24.15 & 13.92 & 8.22 & 33.71 & 28.82 & 30.69 \\
E2T-PTR (proposed) & 10710 & \textbf{42.09} & \textbf{25.13} & \textbf{14.84} & \textbf{8.99} & \textbf{35.86} & \textbf{30.01} & \textbf{32.61} \\
C-SCL~\cite{feng2023aligning} & 14567 & 35.91(\textemdash) & 25.91(\textemdash) & 21.31(\textemdash) & 18.89(\textemdash) & \textemdash & \textemdash & \textemdash \\
C-SCL* & 14407 & 34.87(44.14) & 25.32(31.61) & \textbf{21.17(25.67)} & \textbf{18.98(22.51)} & 36.97 & 34.31 & 35.51 \\
E2T-PTR (proposed) & 14407 & \textbf{34.92(44.31)} & \textbf{25.43(31.67)} & 21.00(25.52) & 18.59(22.22) & \textbf{37.15} & 33.93 & 35.39 \\
EEG2Text*  & 18791 & 58.06 & 49.98 & 46.21 & 44.13 & 52.31 & 48.76 & 50.41 \\
E2T-PTR (proposed) & 18791 & \textbf{59.20} & \textbf{50.77} & \textbf{46.82} & \textbf{44.63} & \textbf{53.76} & \textbf{50.03} & \textbf{51.77} \\
\bottomrule
\end{tabular}%
}
\caption{Comparison of our E2T-PTR framework with previous methods on the ZuCo dataset for three and four reading tasks. * means that our reproduced results. Results enclosed in parentheses are calculated following the approach of EEG2Text, which includes retaining consecutive repeated words in the generated text.}

\label{tab1}
\end{table*}

\begin{table*}[t]
\centering
\resizebox{\textwidth}{!}{
\begin{tabular}{@{}l|l@{}}
\toprule
\multirow{3}{*}{(1)} & Ground Truth: He \textbf{was} first \textit{appointed} to fill the \underline{Senate} \textbf{seat of}\textit{ Ernest Lundeen} who had \textbf{died in} office.\\ \cmidrule(l){2-2} 
 & EEG2Text: \textbf{was} a \textit{elected} to the the \underline{position} \textbf{seat} \textit{in} the \textit{Hemy} in died \textbf{died} \textbf{in} 18 in \\ \cmidrule(l){2-2} 
 & E2T-PTR: \textbf{was} the \textit{elected} to the the \underline{position} \textbf{seat of} \textit{John Hemy}, resigned \underline{resigned} \textbf{in office}. \\ \midrule

\multirow{3}{*}{(2)} & Ground Truth: \underline{Jeb} \textbf{Bush} \textbf{was born in} \underline{Midland}, \textbf{Texas}, where \textbf{his father was} running an oil drill \textbf{company}.\\ \cmidrule(l){2-2} 
 & DeWave: \underline{uan} \textbf{Bush was} a in 18way, \textbf{Texas}, in he \textbf{father was} an insurance refinery \textbf{company}. \\ \cmidrule(l){2-2} 
 & E2T-PTR: \underline{uan} \textbf{Bush}\textbf{ was born in} \underline{Newway}, \textbf{Texas}, and\textbf{ his father was} a a insurance company \textbf{company}. \\ 
 \midrule
 \multirow{3}{*}{(3)} & \begin{tabular}[c]{@{}l@{}}Ground Truth: After Raymond graduated from high \textbf{school}, \textbf{he} enrolled \textbf{in the} \underline{"Universidad del Sagrado Corazon"}  \\  \textbf{(University of the Sacred Heart)} of San Juan, where \textbf{he} earned a \underline{Bachelors} Degree ...\end{tabular} \\ \cmidrule(l){2-2} 
  & \begin{tabular}[c]{@{}l@{}}E2T-PTR: the's from Yale \textbf{school}, \textbf{he} went \textbf{in the} \underline{UniversityAmericancleities de Reyrado Corazon"}  \\ \textbf{(University of the Sacred Heart)} in Spain Francisco, Puerto \textbf{he} studied a \underline{Bachelor.ors} ... \end{tabular} \\ 

\bottomrule
\end{tabular}
}
\caption{EEG-to-Text decoding results. \textbf{Bold} words indicate exact match, \textit{Italic} words indicate semantic resemblance, and \underline{Underline} words indicate error match. We evaluate the translation performance of the same test sentences reported in EEG2Text, DeWave.}
\label{tab2}
\end{table*}

\section{Experiments}
\subsection{Datasets and Evaluation} 
We pre-trained our CET-MAE models under three, four, and five reading tasks in ZuCo v1.0 and ZuCo v2.0. For fairness, we assessed the performance of E2T-PTR for the EEG-to-Text task under the identical dataset scale used during the pre-training phase. We adopt the BLEU and ROUGE-1 scores for evaluating the EEG-to-Text generation performance. More details are presented in Appendix~\ref{appendixb}.\par
\subsection{Implementation Details}
The CET-MAE model features a robust EEG encoder with transformer encoder blocks (6 layers, 2048 hidden dimensions, and 8 attention heads). The EEG decoder is a lightweight transformer encoder of 1 layer with 8 heads. The multi-stream transformer encoder is designed with 1 layer, a 4096 hidden dimension, and 16 attention heads. The mask ratios for EEG feature sequences and textual tokens are set at 75\% (which can achieve the best results based on trial-and-error). For the CET-MAE pertaining objective {$L$}, we set {$\lambda _{T}$}=0.1, {$\lambda _{E}$}=1, {$\lambda _{CL}$}=0.01. This setting is refined through experiments to balance the gradients of each loss in the overall training objective, ensuring that the model learns effectively from each task. We pre-train the CET-MAE model from scratch for 100 epochs. Subsequently, we fine-tune the E2T-PTR model for EEG-to-Text tasks over 50 epochs, employing a batch size of 32 and utilizing the AdamW optimizer. More details are provided in Appendix~\ref{appendixb}.

\begin{table}[t]
\resizebox{0.485\textwidth}{!}{
\begin{tabular}{@{}c|c|llll@{}}
\toprule
\multirow{2}{*}{\begin{tabular}[c]{@{}c@{}}EEG Mask \\ Ratio (\%)\end{tabular}} & \multirow{2}{*}{\begin{tabular}[c]{@{}c@{}}Text Mask   \\ Ratio (\%)\end{tabular}} & \multicolumn{4}{c}{BLEU-N (\%)} \\
 &  & N=1 & N=2 & N=3 & N=4 \\ \midrule
25 & 25 & \textbf{42.14} & 25.02 & 14.55 & 8.62 \\
50 & 25 & 41.74 & 24.75 & 14.39 & 8.52 \\
50 & 50 & 41.80 & 24.69 & 14.25 & 8.40 \\
75 & 50 & 41.93 & 25.02 & 14.72 & 8.81 \\
75 & 75 & 42.09 & \textbf{25.13} & \textbf{14.84} & \textbf{8.99} \\ \bottomrule
\end{tabular}
}
\caption{The performance of our E2T-PTR framework under different combinations of CET-MAE mask ratios rising from 25\% to 50\% , and to 75\% across three reading tasks.}
\label{tab3}
\end{table}

\subsection{Main Results}
Table~\ref{tab1} shows the performance of our E2T-PTR framework on the ZuCo benchmarks. In three reading tasks, E2T-PTR achieves BLEU-1 to BLEU-4 SOTA scores of 42.09\%, 25.13\%, 14.84\%, and 8.99\%, respectively. Moreover, it outperforms best in ROUGE-1 Precision, Recall, and F1 scores compared to recent works. Notably, without removing repetitive generated word tokens, E2T-PTR surpasses C-SCL in BLEU-1 and BLEU-2 scores across four reading tasks. Particularly under the five reading tasks with 18791 training samples, E2T-PTR scores 59.20\%, 50.77\%, 46.82\%, and 44.63\% in BLEU-1 to BLEU-4, significantly exceeding the baseline work EEG2Text.

\begin{table*}[!t]
\centering
\resizebox{\textwidth}{!}{
\begin{tabular}{@{}c|c|cccc|ccc@{}}
\toprule
\multirow{2}{*}{Model} & \multirow{2}{*}{Validation Strategy} & \multicolumn{4}{c|}{BLEU-N(\%)} & \multicolumn{3}{c}{ROUGE-1   (\%)} \\
 &  & N=1 & N=2 & N=3 & N=4 & P & R & F \\ \midrule
\multirow{2}{*}{E2T-PTR} & Split each subject's data in an 8:1:1 ratio & 42.09 & 25.13 & 14.84 & 8.99 & 35.86 & 30.01 & 32.61 \\
 & Leave-one-subject-out Cross-validation & 44.98 & 27.57 & 17.23 & 10.99 & 38.74 & 31.84 & 34.82 \\ \bottomrule
\end{tabular}
}
\caption{Performance comparison of E2T-PTR frameworks between two different data splitting strategies under three reading tasks and used BLEU-N (\%) and ROUGE-1 (\%) as the evaluation metrics.}
\label{tab4}
\end{table*}


\begin{table*}[!t]
\centering
\resizebox{\textwidth}{!}{
\begin{tabular}{@{}c|c|c|c|cccc|ccc}
\toprule
\multirow{2}{*}{\begin{tabular}[c]{@{}c@{}}Sentence-level  EEG \\ feature   sequences\end{tabular}} & \multirow{2}{*}{CET-MAE} & \multirow{2}{*}{E2T-PTR} & \multirow{2}{*}{\begin{tabular}[c]{@{}c@{}}Training  \\ Sample\end{tabular}} & \multicolumn{4}{c|}{BLEU-N (\%)} & \multicolumn{3}{c}{ROUGE-1 (\%)} \\ 
 &  &  &  & N=1 & N=2 & N=3 & N=4 & P & R & F \\ \midrule
 \multicolumn{1}{c|}{\ding{53}} & \multicolumn{1}{c|}{\ding{53}} & \multicolumn{1}{c|}{\ding{53}} & 10710 & 41.16 & 23.99 & 13.49 & 7.68 & 34.68 & 28.96 & 31.45  \\
\multicolumn{1}{c|}{\checkmark} & \multicolumn{1}{c|}{\ding{53}} & \multicolumn{1}{c|}{\ding{53}} & 10710 & 41.63 & 24.48 & 13.96 & 8.06 & 35.13 & 29.27 & 31.83 \\
\multicolumn{1}{c|}{\checkmark} & \multicolumn{1}{c|}{\checkmark} & \multicolumn{1}{c|}{\ding{53}}  & 10710 & 41.88 & 24.85 & 14.52 & 8.74 & 35.26 & 29.50 & 32.02 \\
\multicolumn{1}{c|}{\checkmark}  & \multicolumn{1}{c|}{\checkmark}  & \multicolumn{1}{c|}{\checkmark}  & 10710 & \textbf{42.09} & \textbf{25.13} & \textbf{14.84} & \textbf{8.99} & \textbf{35.86} & \textbf{30.01} & \textbf{32.61} \\ \bottomrule
\end{tabular}
}
\caption{The results of ablation experiments on CET-MAE and E2T-PTR structures under three reading tasks. We verified the effectiveness of each component and used BLEU-N (\%) and ROUGE-1 (\%) as the evaluation metrics.}
\label{tab5}
\end{table*}

Table\ref{tab2} presents a comparative analysis of the decoding results between our model and other models under three reading tasks. Our model E2T-PTR demonstrates an enhanced ability to generate more complete grammatical structures, which is evident from the reduced error rates and increased semantic coherence in the decoded sentences, exemplified by expressions such as “\textbf{his father was}” and “\textbf{Bush was born in}”. Our model also excels in decoding common and proper nouns, such as “\textbf{office}” and “\textbf{University of the Sacred Heart}”. It also adeptly produces semantically similar words, such as, “\textit{appointed}” vs “\textit{elected}”, and “\textit{Ernest Lundeen}” vs “\textit{John Hemy}”. Intriguingly, upon expanding our training samples to 1.75 times (10710 to 18791), we observe an obvious improvement in the translation quality of the model, especially concerning fine-grained recognition. More comprehensive results are included in the Appendix~\ref{appendixd}. 
\par

Our investigation delved into the transfer performance of CET-MAE across varying EEG and text masking ratios under three reading tasks. Table \ref{tab3} details the performance shifts under different combinations of masking ratios rising from 25\% to 50\%, and to 75\%.  We discovered that the CET-MAE model excels at the higher masking ratios of 75\%, starkly contrasting with the traditional 15\% mask ratio suggested in BERT. This result is consistent with recent findings in multi-modal masked models~\cite{ma2022simvtp,geng2022multimodal}, suggesting that inter-modal interactions may promote performance improvement. We further ponder this phenomenon and suggest that, in terms of CET-MAE structure, it appears to be suited for reconstructing masked EEG features and predicting masked word tokens. In terms of the masking strategy, forcefully masking sentence-level EEG embeddings can better compel the model to learn global semantic information. Furthermore, we discuss the overall masking ratio for the EEG, the natural EEG masking ratio under three reading tasks is 32.51\% as mentioned in Appendix \ref{appendixa}. Therefore, the total masking ratio for the EEG is 83.13\% \footnote{Overall Masking Ratio = NMR +  (1 - NMR) × CET-MAE Masking Ratio.} (32.51\% of natural + 50.62\% of CET-MAE masked). \par

For a more rigorous validation,  we further implemented the leave-one-subject-out validation strategy for both the CET-MAE model and the E2T-PTR framework, detailed in Table~\ref{tab4}.  This validation approach proved extremely valuable in testing the generalization performance across different subjects within the EEG dataset. Given the inherent noise and individual variability in EEG data, it is crucial to evaluate how well a model performs under such conditions. The results obtained from the leave-one-subject-out validation not only exceeded our initial performance metrics presented in Table~\ref{tab1} but also underscored the strong generalizability of our models. These results affirm the ability of our models to effectively manage the inherent variability in EEG data, thereby demonstrating robust performance as each subject’s data was sequentially excluded from the training set.

\begin{table}[!t]
\begin{tabular}{@{}lccc@{}}
\toprule
\multirow{2}{*}{Metrics (\%)} & \multicolumn{3}{c}{Our SSL Models} \\ \cmidrule(l){2-4} 
 & CET & ET-MAE & CET-MAE \\ \midrule
BLEU-1 & 41.77 & 41.80 & \textbf{42.09} \\
BLEU-2 & 24.68 & 24.72 & \textbf{25.13} \\
BLEU-3 & 14.33 & 14.43 & \textbf{14.84} \\
BLEU-4 & 8.60 & 8.53 & \textbf{8.99} \\
ROUGE-1 P & 35.59 & 35.06 & \textbf{35.86} \\
ROUGE-1 R & \textbf{30.11} & 29.31 & 30.01 \\
ROUGE-1 F & 32.51 & 31.82 & \textbf{32.61} \\ \bottomrule
\end{tabular}
\caption{Evaluating transfer performance across CET, ET-MAE, and CET-MAE under three reading tasks.}
\label{tab6}
\vspace{-7pt}
\end{table}

\subsection{Ablation Studies}
Table \ref{tab5} details the ablation experiments, affirming the effectiveness of each component in our approaches for EEG-to-Text generation quality.
First, sentence-level EEG features positively impact BLEU scores, notably BLEU-1, underscoring their importance in capturing essential semantic information for improved text generation.
Second, CET-MAE, focusing on masked signal modeling and contrastive learning between EEG and text, is fundamental. Integrating CET-MAE with the baseline framework~\cite{wang2022open} significantly boosts BLEU scores, especially BLEU-4.
Third, combining E2T-PTR with CET-MAE enhances performance across metrics, particularly Precision, Recall, and F1 score of ROUGE-1, showcasing E2T-PTR's role in effectively transferring CET-MAE's learned representations.


\subsection{Transfer Performance of SSL Models}
We further pre-train and compare the transfer performance of the following SSL models: 1) Contrastive EEG-Text (CET) learning model: The CET that has no reconstruction objective. For a fair comparison, we implement CET using the same encoder architecture (modal-specific encoders + multi-stream encoder) with CET-MAE but remove the reconstruction task ({$L_{E}$} and {$L_{T}$} ). We use this model to investigate the impact of contrastive learning. 2) EEG-text masked autoencoder (ET-MAE) model: The ET-MAE has the same architecture as CET-MAE but the contrastive loss ({$L_{CL}$}) is set to 0. The masking strategy is the same as CET-MAE.  We use this model to examine the effectiveness of masked signal modeling. 
3) Our proposed CET-MAE is detailed in Section \ref{sec:methods}.

To ensure fairness, CET and ET-MAE are pre-trained with the same pipeline as CET-MAE. We assess their EEG-to-Text transfer performance using the E2T-PTR framework.
Results in Table \ref{tab6} demonstrate CET-MAE's superiority over two other SSL models (CET and ET-MAE) across most evaluation metrics. Specifically, CET-MAE achieves improvements of 0.32\%, 0.45\%, 0.51\%, and 0.39\% in BLEU-1 to BLEU-4, respectively, compared to CET. Against ET-MAE, CET-MAE records increases of 0.29\%, 0.41\%, 0.41\%, and 0.46\% for these metrics, respectively.
The trend of enhancement is consistent in ROUGE-1 metrics as well.



\section{Conclusion}

This study contributes to the development of EEG-based language decoding by introducing an effective EEG-text pre-trained model, CET-MAE, and a highly capable and LLM-empowered EEG-to-Text decoding framework, E2T-PTR.
CET-MAE uses a multi-stream architecture to incorporate both intra- and cross-modality SSL within one unified system:
1) Intra-modality streams explore representative embeddings that reflect the intrinsic characteristics of EEG or text sequences, leveraging masked modeling with a mask ratio of up to 75\%;
2) Inter-modality stream provides dual-modal representations to enhance intra-modality reconstruction and constrains the encoder to maximize semantic consistency between text and its corresponding EEG sequences.
E2T-PTR transfers pre-trained EEG representations and leverages BART's capabilities for text generation from these consistent and representative features. 
Extensive experiments on the latest text-evoked EEG dataset, ZuCo, demonstrate the superiority of this work in both qualitative and quantitative assessments. 
The proposed CET-MAE model shows great potential for enhancing EEG-based language decoding tasks and could be utilized for other inner speech BCI datasets.


\par


\clearpage
\section*{Limitation}
The limitations of our study are summarized as follows:
\par
\textbf{Dataset Scale:} The performance of both the CET-MAE model and the E2T-PTR framework is constrained by the scale of currently available datasets. We are in the process of developing our datasets to fully exploit the potential of our models and frameworks.  \par

\textbf{Teacher Forcing:} 
While our results are pushing the open vocabulary EEG-to-Text decoding performances to a new SOTA, they still depend on the implicit use of teacher forcing, a common precondition in recent studiess~\cite{wang2022open,duan2024dewave,feng2023aligning,xi2023unicorn}. This reliance on teacher forcing could be constraining the full capabilities of the LLMs.  Noted that recent work~\cite{yang2024decode} has reported promising results with the autoregressive capabilities of large speech models like Whisper~\cite{radford2023robust} on the MEG datasets~\cite{schoffelen2019204}. This may offer potential solutions to the challenges of using teacher forcing in the EEG-to-Text field. Our future work will aim to verify the correctness of the aforementioned new methods and explore the autoregressive capabilities of LLMs to reduce reliance on teacher forcing. \par 


\textbf{Exploration of LLMs:}
We plan to explore more advanced LLMs to enhance our EEG-to-Text decoding capabilities. This will involve testing new models and techniques to improve performances and uncover deeper insights from EEG data. \par

\section*{Ethics Statement}
In this work, we do not generate new EEG data, nor do we perform experiments on human subjects. We use the publicly available ZuCo v1.0 and ZuCo v2.0 datasets without any restrictions. We do not anticipate any harmful applications of our work.

\section*{Acknowledgements}
This work is supported by the National Natural Science Foundation of China (No. 62306089) and the China Post-doctoral Science Foundation (Nos. 2023M730873 and GZB20230960).

\vspace{-10pt}
\bibliography{custom}

\clearpage

\appendix

\section{Natural Masking Ratio of Datasets}
\label{appendixa}
To provide a clear perspective, we present the detailed statistics of the NMR of EEG feature sequences for three categories of reading task combinations in Table\ref{tab7}.

\section{Datasets and Implementation Details}
\label{appendixb}
We utilize the combination of both ZuCo v1.0 and ZuCo v2.0 to form the final ZuCo benchmark. The EEG features are collected with a 128-channel system under the sampling rate of 500Hz. After the noise canceling process, only 105 channels are used. There are 8 frequency bands determined in the ZuCo dataset as follows: theta1 (4–6 Hz), theta2 (6.5–8 Hz) alpha1 (8.5–10 Hz), alpha2 (10.5–13 Hz), beta1 (13.5–18 Hz) beta2 (18.5–30 Hz) and gamma1 (30.5–40 Hz) and gamma2 (40–49.5 Hz). The Hilbert transform is applied in each of these time series. 
The final features of the EEG are formed by concatenating features from all 8 frequency bands, resulting in a vector with a dimension of 840. For three reading tasks, we pre-train and fine-tune the models on “\textit{SR v1.0 + NR v1.0 + NR v2.0}”. For four reading tasks, we choose the combination of “\textit{SR v1.0 + NR v1.0 + NR v2.0 + TSR v1.0}”. For five reading tasks, the models are pre-trained and fine-tuned on “\textit{SR v1.0 + NR v1.0 + NR v2.0 + TSR v1.0 + TSR v2.0}”. During pre-training, the datasets were split into training and testing sets in a 90\% to 10\% ratio. During the EEG-to-Text fine-tuning phase, the datasets were further divided into training, validation, and testing sets with an 80\%, 10\%, and 10\% split respectively. The test set samples remained consistent throughout the above two stages. The dataset statistics of EEG-to-Text decoding are detailed in Table~\ref{tab8}. 
Our training hyper-parameters are listed in Table~\ref{tab9}. To ensure a fair comparison, we conducted both pre-training and fine-tuning for the EEG-to-Text decoding task using datasets with the same combinations of reading tasks. 


\section{Generated Samples}
\label{appendixd}
We show more details in EEG-to-Text translation results generated on our models  in Table~\ref{tab10}, Table~\ref{tab11}, and Table\ref{tab12}.  In our experiments, we aim to select the same sentences from the test sets of three, four, and five reading tasks where feasible. This enables us to directly observe and compare the generated results with the ground truth across different task conditions.

\section{Subject-independent Performance}
\label{appendixf}
\begin{figure*}[!ht]
    \centering
    \includegraphics[width=1.0\textwidth]{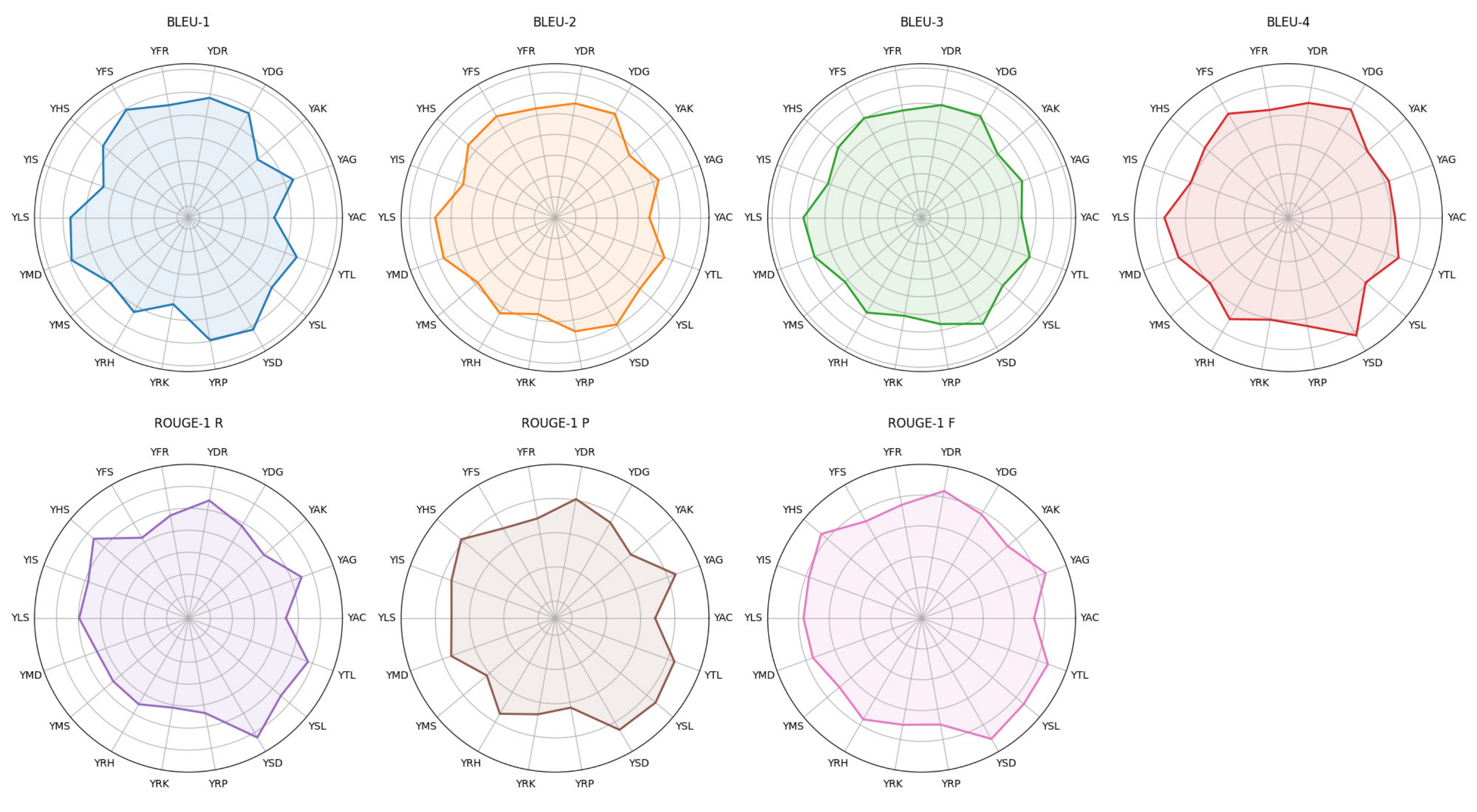} 
    \caption{The radar chart of 18 subjects from Subject YAG to YSD on each metric.}
    \label{Figure3}
\end{figure*}

\begin{figure*}[!ht]
    \centering
    \includegraphics[width=1.0\textwidth]{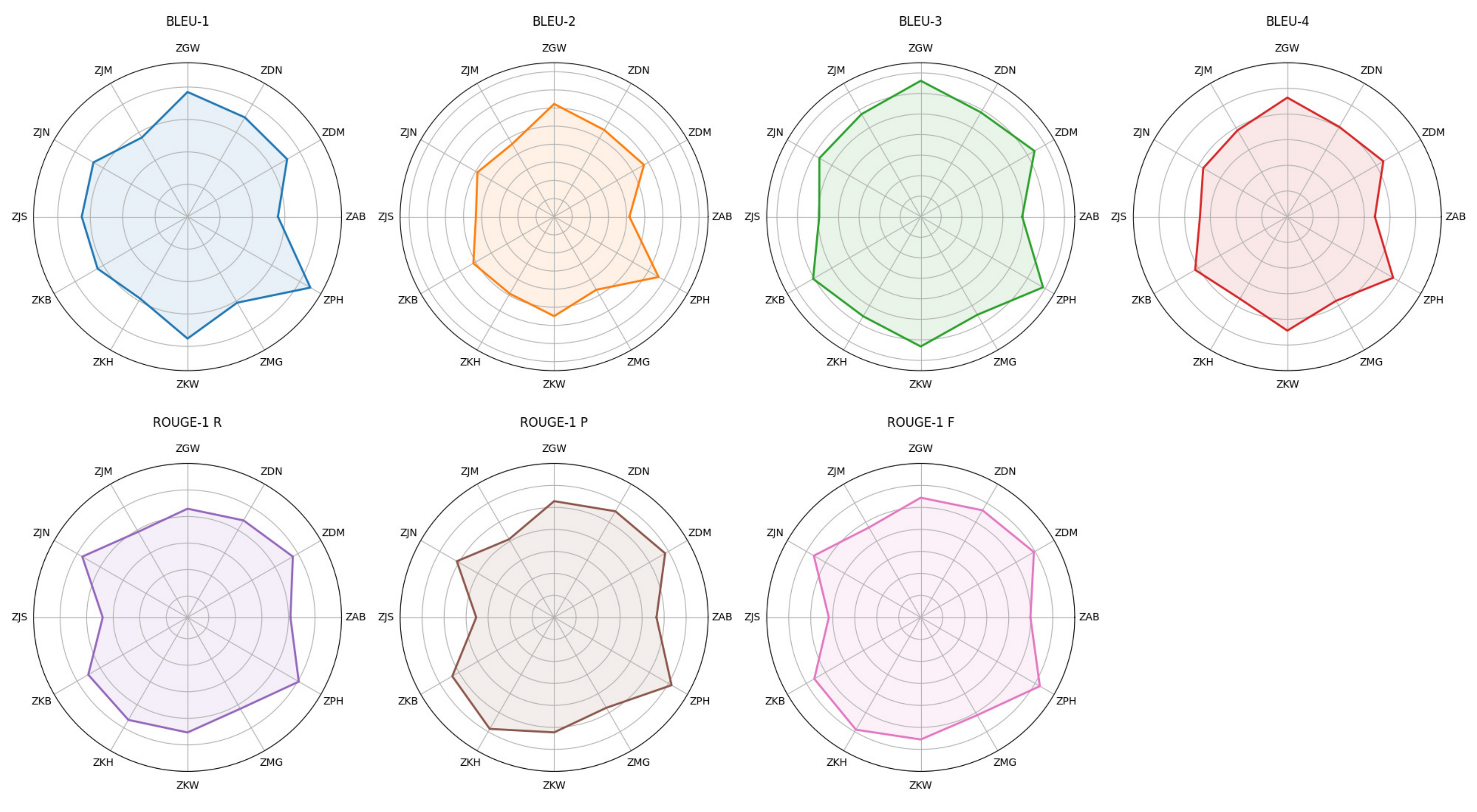} 
    \caption{The radar chart of 12 subjects from Subject ZKW-ZJS on each metric.}
    \label{Figure4}
\end{figure*}

As reported in Table\ref{tab1}, we present the average BLEU-N and ROUGE-1 scores for all 30 subjects. However, considering the individual variations of brain activities during semantic processing and cognitive operations within different subjects, we further provide individual BLEU-N and ROUGE-1 scores for each subject. We use radar charts shown in Figure\ref{Figure3} and Figure\ref{Figure4} to visually represent these differences, allowing for an intuitive comparison across subjects. For a detailed numeric breakdown of these variances, refer to Table\ref{tab13} and Table\ref{tab14}.

\section{Impact of the Masking Strategy}
\label{appendixg}


The masking strategy is crucial in Masked Autoencoders. For the text, the BERT masking strategy has proven highly effective. For the EEG modality, we introduce a pivotal design that involves mandatory masking of sentence-level EEG feature sequences, as detailed in Section \ref{sec:intro1}. We delve into the impact of this strategy on the EEG-to-Text decoding task. 
Comparative results between random and forced masking strategies are presented in Table \ref{tab15}. The forced masking strategy outperforms the random masking strategy in the EEG-to-Text decoding, highlighting the efficacy of our proposed strategy in compelling the model to reconstruct the contextual semantics within sentence-level EEG feature sequences comprehensively.

\section{Impact of the Multi-Stream Design}
\label{appendixh}
Our investigation, as detailed in Table \ref{tab16}, reveals the transfer performance of a multi-stream design in the CET-MAE and E2T-PTR frameworks.  The multi-stream approach, which provides the specialized handling of text and EEG using separate streams, outperformed a single joint stream design. Notably, in the E2T-PTR framework, leveraging the EEG-specific stream for fine-tuning yielded a marked improvement in EEG-to-Text task performance over a joint modality stream. This modality-focused approach appears to capitalize on the nuanced semantic information inherent in EEG embeddings, resulting in a more sophisticated and contextually relevant latent space. This is substantiated by the observed uptick in BLEU and ROUGE metrics. Our study underscores the criticality of fine-grained, modality-specific processing approaches in the domain of EEG-Text representation learning.

\begin{table*}[t]
\centering
\resizebox{1.0\textwidth}{!}{
\begin{tabular}{@{}l|c|c|c@{}}
\toprule
Reading Tasks & Missing Paris & Total word tokens & NMR(\%) \\ \midrule
SR   v1.0 + NR v1.0+NR   v2.0 & 90362 & 277966 & 32.51 \\
SR   v1.0 + NR v1.0+NR   v2.0+TSR v1.0 & 137460 & 373817 & 36.77 \\
SR   v1.0 + NR v1.0+NR   v2.0+TSR v1.0+TSR v2.0 & 204089 & 515979 & 39.55 \\ \bottomrule
\end{tabular}
}
\caption{Statistics for natural masking ratios under three, four, and five reading tasks in ZuCo benchmarks.}
\label{tab7}
\end{table*}

\begin{table*}[t]
\centering
\resizebox{1.0\textwidth}{!}{
\begin{tabular}{@{}l|c|c|c@{}}
\toprule
\begin{tabular}[c]{@{}l@{}}Reading  Task\end{tabular} & 
\begin{tabular}[c]{@{}c@{}}Training Sample\end{tabular} &
\begin{tabular}[c]{@{}c@{}}Validation Sample\end{tabular} & 
\begin{tabular}[c]{@{}c@{}}Testing Sample\end{tabular} \\ \midrule
SR v1.0 + NR v1.0+NR v2.0 & 10710 & 1332 & 1407 \\ 
SRv1.0+NRv1.0+NRv2.0+TSRv1.0 & 14407 & 1790 & 1799 \\ 
SRv1.0+NRv1.0+NRv2.0+TSRv1.0+TSRv2.0 & 18791 & 2287 & 2404 \\ \bottomrule
\end{tabular}}
\caption{Dataset Statistics of the EEG-to-Text decoding. SR: Normal Reading (Sentiment), NR: Normal Reading (Wikipedia), TSR: Task Specific Reading (Wikipedia). }
\label{tab8}
\end{table*}

\begin{table*}[!t]
\centering
\begin{tabularx}{1.0\textwidth}{@{}l|YYYYYY@{}}
\toprule
\textbf{Hyperparameters} & \multicolumn{3}{c}{\textbf{Pre-training}} & \multicolumn{3}{c}{\textbf{Fine-tuning}} \\ \midrule
Models & \multicolumn{3}{c}{CET-MAE} & \multicolumn{3}{c}{E2T-PTR} \\
Reading Tasks & 3 & 4 & 5 & 3 & 4 & 5 \\
Datasets Splits & \multicolumn{3}{c}{9:1} & \multicolumn{3}{c}{8:1:1} \\
Epochs & \multicolumn{3}{c}{100} & 50 & 40 & 40 \\
Batch Size & \multicolumn{3}{c}{32} & \multicolumn{3}{c}{32} \\
Learning Rate & \multicolumn{3}{c}{5e-7} & 2e-7 & 2e-5 & 2e-5 \\
Optimizer & \multicolumn{6}{c}{AdamW, weight decay= 1e-2, betas =(0.9,0.999)} \\
LR Scheduler & \multicolumn{6}{c}{Cosine Annealing, T\_max=20} \\
GPUs & \multicolumn{6}{c}{RTX4090} \\ \bottomrule
\end{tabularx}
\caption{Implementation details in our pre-training and fine-tuning.}
\label{tab9}
\end{table*}

\begin{table*}[t]
\centering
\resizebox{1.0\textwidth}{!}{
\begin{tabular}{@{}c|l@{}}
\toprule
\multirow{2}{*}{(1)} & \begin{tabular}[c]{@{}l@{}}Ground Truth: At the urging of \textbf{his wife}, Columba, a devout Mexican \textbf{Catholic}, the Protestant Bush became a Roman Catholic.
 \end{tabular} \\ \cmidrule(l){2-2} 
 
 & \begin{tabular}[c]{@{}l@{}}E2T-PTR: the time of \textbf{his wife}, hea, he former Catholic \textbf{Catholic}, he actor pastorman a Catholic \textbf{Catholic}. \end{tabular} \\ \midrule
 
 \multirow{2}{*}{(2)} & \begin{tabular}[c]{@{}l@{}}Ground Truth: While attending a motorcycle race, \textbf{he} met a local girl \textbf{named} \underline{Columba Garnica Gallo}, \textit{whom} \textbf{he} eventually \textbf{married}. \end{tabular} \\ \cmidrule(l){2-2} 
 
 & \begin{tabular}[c]{@{}l@{}}E2T-PTR: in the local school, \textbf{he} was his man boy \textbf{named} \underline{Marya,ett,o}, \textit{who} \textbf{he} later \textbf{married}. \end{tabular} \\ \midrule
 
 \multirow{2}{*}{(3)} & \begin{tabular}[c]{@{}l@{}}Ground Truth: He then enrolled at Phillips Andover, a \textbf{private} boarding \textbf{school in Massachusetts} already attended \textbf{by his} \\ \underline{brother George}.  \end{tabular} \\ \cmidrule(l){2-2} 
 
 & \begin{tabular}[c]{@{}l@{}}E2T-PTR: was went in the Academy Mary College where \textbf{private} school \textbf{school in Massachusetts}. known \textbf{by his} \underline{father},. \end{tabular} \\ \midrule
 
\multirow{2}{*}{(4)} & \begin{tabular}[c]{@{}l@{}}Ground Truth: He took \textbf{a job} in real \textbf{estate} with Armando Codina, a \underline{32}\textbf{-year-old} Cuban \textbf{immigrant} and \underline{self}\textbf{-made} American \\ \textbf{millionaire.} \end{tabular} \\ \cmidrule(l){2-2} 

 & \begin{tabular}[c]{@{}l@{}}E2T-PTR: was \textbf{a job} as the \textbf{estate} in theando Iice in who \underline{company}\textbf{-year-old} Italian \textbf{immigrant}. \underline{former}\textbf{-made} millionaire \\ \textbf{millionaire}. \end{tabular} \\ \midrule
 
\multirow{2}{*}{(5)} & \begin{tabular}[c]{@{}l@{}}Ground Truth: After earning \textbf{his} \textit{degree}, Bush went \textbf{to work} in an entry level \textbf{position} in the international division of Texas  \\ Commerce Bank, which was run \textbf{by} \underline{Ben Love}.  \end{tabular} \\ \cmidrule(l){2-2} 

 & \begin{tabular}[c]{@{}l@{}}E2T-PTR: the \textbf{his} \textit{bachelor} in he became \textbf{to work} for the office- \textbf{position} at the Department banking of the Instruments..\\ where was later \textbf{by} \underline{theitott} \end{tabular} \\ \midrule
 
 \multirow{2}{*}{(6)} & \begin{tabular}[c]{@{}l@{}}Ground Truth: He later \textbf{became} \textit{an} \underline{educator}, teaching music theory at \textbf{the University of} \underline{the District} \textbf{of Columbia}; he was \textbf{also} \\ director \textbf{of the} District \textbf{of} \textit{Columbia} \underline{Music Center jazz workshop band}.
  \end{tabular} \\ \cmidrule(l){2-2} 
  
 & \begin{tabular}[c]{@{}l@{}}E2T-PTR: was \textbf{became} \textit{a} \underline{American} and and at and and \textbf{the University of} \underline{California West} \textbf{of Columbia}. and also \textbf{also} \\ a \textbf{of the} school \textbf{of} \textit{Columbia's} \underline{Department. department..}\end{tabular} \\ \midrule
 
  \multirow{2}{*}{(7)} & \begin{tabular}[c]{@{}l@{}}Ground Truth: Bush stayed in Houston with another family to finish the school \textbf{year}, \textbf{and} spent most \textbf{summers}  and holidays \\ at \textbf{the} \textit{family} estate, known as \textbf{the Bush Compound.}
  \end{tabular} \\ \cmidrule(l){2-2} 
  
 & \begin{tabular}[c]{@{}l@{}}E2T-PTR: was in the until his family, raise his year \textbf{year}. \textbf{and} then the of  in \textbf{summers} there \textbf{the} \textit{family's}. including \\  as \textbf{the Bush Ranchound}. \end{tabular} \\ \midrule
 
\multirow{2}{*}{(8)} & \begin{tabular}[c]{@{}l@{}}Ground Truth: Robert Henry Dee (\textbf{born} \underline{May 18, 1933 in Quincy,} \textbf{Massachusetts}) \textbf{is a} former three-sport letterman  at Holy \textbf{Cross} \\ \textbf{College}  who was one \textbf{of the} first \textit{players} signed \textbf{by the} \underline{Boston Patriots in 1960}.
 \end{tabular} \\ \cmidrule(l){2 -2} 
 
& \begin{tabular}[c]{@{}l@{}}E2T-PTR: Frost, (\textbf{born} \underline{April 5, 18) New}, \textbf{Massachusetts}) \textbf{is a} retired United-timeport star carrier and the \textbf{Cross College}. \textit{played} a \textbf{of} \\  \textbf{the} founders African to \textbf{by the} \underline{University Celtics. the}. \end{tabular} \\ \bottomrule
\end{tabular}
}
\caption{EEG-to-Text decoding example results on test sentences under three reading tasks. \textbf{Bold} words indicate exact match, \textit{Italic} words indicate semantic resemblance, and \underline{Underline} words indicate error match.}
\label{tab10}
\end{table*}

\begin{table*}[!th]
\vspace{-1.5cm}
\centering
\resizebox{\textwidth}{!}{
\begin{tabular}{@{}c|l@{}}
\toprule
\multirow{2}{*}{(1)} & \begin{tabular}[c]{@{}l@{}}Ground Truth: At the urging \textbf{of his} \underline{wife}, Columba, a devout Mexican \textbf{Catholic}, the Protestant Bush became a Roman \textbf{Catholic}. \end{tabular} \\ \cmidrule(l){2-2} 
 
 & \begin{tabular}[c]{@{}l@{}}E2T-PTR: the academy \textbf{of his} \underline{mother}, hea, she young \textbf{Catholic}\underline{-}, she young preacher co an Catholic \textbf{Catholic} \underline{in} \end{tabular} \\ \midrule
 
 \multirow{2}{*}{(2)} & \begin{tabular}[c]{@{}l@{}}Ground Truth: While attending a motorcycle race, \textbf{he met a} local girl \textbf{named} \underline{Columba Garnica Gallo}, \textit{whom} \textbf{he} \underline{eventually} \textbf{married}. \end{tabular} \\ \cmidrule(l){2-2} 
 
 & \begin{tabular}[c]{@{}l@{}}E2T-PTR: serving the Louisiana school he \textbf{he met a} man hero \textbf{named} \underline{Dela Jacksonett.ienne}. \textit{who} \textbf{he} \underline{would struck}. \end{tabular} \\ \midrule
 
 \multirow{2}{*}{(3)} & \begin{tabular}[c]{@{}l@{}}Ground Truth: He then enrolled at Phillips Andover, a private boarding school in \textbf{Massachusetts} already attended \textbf{by his} \\ \underline{brother George}.  \end{tabular} \\ \cmidrule(l){2-2} 
 
 & \begin{tabular}[c]{@{}l@{}}E2T-PTR: was returned in the University Mary College \textbf{Massachusetts} public school school \textbf{in} the. owned \textbf{by his} \underline{father},. \end{tabular} \\ \midrule
 
\multirow{2}{*}{(4)} & \begin{tabular}[c]{@{}l@{}}Ground Truth: He took a job in real \textbf{estate with} \underline{Armando Codina}, a \underline{32}\textbf{-year-old} Cuban immigrant and self-made American 
\\ \textbf{millionaire}. \end{tabular} \\ \cmidrule(l){2-2} 

 & \begin{tabular}[c]{@{}l@{}}E2T-PTR: was many second as the \textbf{estate with} \underline{theando Feric}, where \underline{local}\textbf{-year-old} hotel shipping who hotel-trained millionaire \\ \textbf{millionaire} \underline{who} \end{tabular} \\ \midrule
 
\multirow{2}{*}{(5)} & \begin{tabular}[c]{@{}l@{}}Ground Truth: After earning \textbf{his} degree, Bush \textbf{went to work} in an entry level position in the international division of Texas  \\ Commerce Bank, which was run \textbf{by} \underline{Ben Love}.  \end{tabular} \\ \cmidrule(l){2-2} 

 & \begin{tabular}[c]{@{}l@{}}E2T-PTR: a \textbf{his} Ph at he \textbf{went to work} for the apprentice- role at the Springfield trade of the Instruments. at working he \\ subsequently \textbf{by} \underline{Jamesoittt} \end{tabular} \\ \midrule
 
 \multirow{2}{*}{(6)} & \begin{tabular}[c]{@{}l@{}}Ground Truth: He later became an educator, teaching music theory \textbf{at the University of} \underline{the District} \textbf{of Columbia}; he was \textbf{also} \\ director \textbf{of the} District of \textbf{Columbia} \underline{Music Center jazz workshop band}.
  \end{tabular} \\ \cmidrule(l){2-2} 
  
 & \begin{tabular}[c]{@{}l@{}}E2T-PTR: was earned president assistant at and English at \textbf{at the University of} \underline{Wisconsin Arts} \textbf{of Columbia}, and \textbf{also} the a \\ \textbf{of the} Special School \textbf{Columbia} \underline{Library Project. line..} \end{tabular} \\ \midrule
 
  \multirow{2}{*}{(7)} & \begin{tabular}[c]{@{}l@{}}Ground Truth: Bush stayed \textbf{in} Houston with another family to finish the school \textbf{year}, \textbf{and} spent most \textbf{summers and holidays}  \\ \textbf{at the} \underline{family} \textbf{estate}, known \textbf{as the Bush Compound}.
  \end{tabular} \\ \cmidrule(l){2-2} 
  
 & \begin{tabular}[c]{@{}l@{}}E2T-PTR: was \textbf{in} Hollywood for his oil, work his term \textbf{year}, \textbf{and} to the \textbf{summers and holidays at the} \underline{sprawling} \textbf{estate}, \\ the \textbf{as the Bush Compound}. \end{tabular} \\ \midrule
 
\multirow{2}{*}{(8)} & \begin{tabular}[c]{@{}l@{}}Ground Truth: Robert Henry Dee (born May 18, 1933 in Quincy, \textbf{Massachusetts}) \textbf{is} a \textbf{former} three-sport letterman  at Holy \textit{Cross} \\ \textit{College} \textbf{who} was one of the first players signed by \textbf{the} \underline{Boston Patriots in 1960}.
 \end{tabular} \\ \cmidrule(l){2 -2} 
 
& \begin{tabular}[c]{@{}l@{}}E2T-PTR: Joseph Bol,born July 22, 1923) Ball, \textbf{Massachusetts}) \textbf{is} best \textbf{former} Republican-timeides quarterbackman \textbf{who} the \textit{Cross} \\  \textit{College}, is elected of the founder " to to \textbf{the} \underline{University Bruins. 1993}. \end{tabular} \\ \bottomrule
\end{tabular}
}
\caption{EEG-to-Text decoding example results on test sentences under four reading tasks. \textbf{Bold} words indicate exact match, \textit{Italic} words indicate semantic resemblance, and \underline{Underline} words indicate error match.}
\label{tab11}
\end{table*}

\newpage
\begin{table*}[!t]
\vspace{-0.1cm}
\centering
\resizebox{\textwidth}{!}{
\begin{tabular}{@{}c|l@{}}
\toprule
\multirow{2}{*}{(1)} & \begin{tabular}[c]{@{}l@{}}Ground Truth: At the urging \textbf{of his} \underline{wife}, Columba, a devout Mexican \textbf{Catholic}, the Protestant Bush became a Roman \textbf{Catholic}. \end{tabular} \\ \cmidrule(l){2-2} 

& \begin{tabular}[c]{@{}l@{}}E2T-PTR: the academy \textbf{of his} \underline{mother}, hea, she young Catholic \textbf{Catholic}, she young and accepted a Catholic \textbf{Catholic} \underline{in} \end{tabular} \\ \midrule

\multirow{2}{*}{(2)} & \begin{tabular}[c]{@{}l@{}}Ground Truth: \underline{While attending} \textbf{a motorcycle race, he met a local girl named Columba Garnica Gallo, whom he eventually married.} \end{tabular} \\ \cmidrule(l){2-2} 

& \begin{tabular}[c]{@{}l@{}}E2T-PTR: \underline{serving} \textbf{a motorcycle race, he met a local girl named Columba Garnica Gallo, whom he eventually married.} \end{tabular} \\ \midrule

\multirow{2}{*}{(3)} & \begin{tabular}[c]{@{}l@{}}Ground Truth: \underline{He then} \textbf{enrolled at Phillips Andover, a private boarding school in Massachusetts already attended by his} \\ \textbf{brother George}.  \end{tabular} \\ \cmidrule(l){2-2} 

& \begin{tabular}[c]{@{}l@{}}E2T-PTR: \underline{was} \textbf{enrolled at Phillips Andover, a private boarding school in Massachusetts already attended by his brother George}. \end{tabular} \\ \midrule
 
\multirow{2}{*}{(4)} & \begin{tabular}[c]{@{}l@{}}Ground Truth: He took a \textbf{job} in real \textbf{estate} \textbf{with} \underline{Armando Codina}, a \underline{32}\textbf{-year-old} Cuban immigrant and self-made American 
\\ \textbf{millionaire}. \end{tabular} \\ \cmidrule(l){2-2} 

& \begin{tabular}[c]{@{}l@{}}E2T-PTR: was his \textbf{job} with the \textbf{estate with} \underline{theco Ferela} and and \underline{firm}\textbf{-year-old} firm shipping who hotel-trained \textbf{millionaire} \underline{merchant}. \end{tabular} \\ \midrule
 
\multirow{2}{*}{(5)} & \begin{tabular}[c]{@{}l@{}}Ground Truth: \underline{After earning} \textbf{his degree, Bush went to work in an entry level position in the international division of Texas } \\ \textbf{Commerce Bank, which was run by Ben Love}.  \end{tabular} \\ \cmidrule(l){2-2} 

& \begin{tabular}[c]{@{}l@{}}E2T-PTR: \underline{a} \textbf{his degree, Bush went to work in an entry level position in the international division of Texas Commerce Bank}, \\ \textbf{which was run by Ben Love}. \end{tabular} \\ \midrule
 
\multirow{2}{*}{(6)} & \begin{tabular}[c]{@{}l@{}}Ground Truth: \underline{He later} \textbf{became} \underline{an} \textbf{educator, teaching music theory at the University of the District of Columbia; he was also} \\ \textbf{director of  the District of Columbia Music Center jazz workshop band}.
\end{tabular} \\ \cmidrule(l){2-2} 
  
& \begin{tabular}[c]{@{}l@{}}E2T-PTR: \underline{was} \textbf{became} \underline{president} \textbf{educator, teaching music theory at the University of the District of Columbia; he was also}  \\ \textbf{director of the District of Columbia Music Center jazz workshop band}. \end{tabular} \\ \midrule
 
\multirow{2}{*}{(7)} & \begin{tabular}[c]{@{}l@{}}Ground Truth: Bush stayed \textbf{in} Houston with another family to finish the school \textbf{year}, and \underline{spent} most \textbf{summers and holidays}  \\ \textbf{at the} \underline{family} \textbf{estate}, known \textbf{as the Bush Compound}.
\end{tabular} \\ \cmidrule(l){2-2} 
  
 & \begin{tabular}[c]{@{}l@{}}E2T-PTR: is \textbf{in} Hollywood for his company, work his war \textbf{year}. \textbf{and} \underline{enrolled} the \textbf{summers and holidays at the} \underline{sprawling} \textbf{estate}, \\ the \textbf{as the Bush Compound}. \end{tabular} \\ \midrule
 
\multirow{2}{*}{(8)} & \begin{tabular}[c]{@{}l@{}}Ground Truth: \underline{Robert Henry} \textbf{Dee (born May 18, 1933 in Quincy, Massachusetts) is a former three-sport letterman  at Holy Cross} \\ \textbf{College who was one of the first players signed by the Boston Patriots in 1960}.
 \end{tabular} \\ \cmidrule(l){2 -2} 
 
& \begin{tabular}[c]{@{}l@{}}E2T-PTR: \underline{Emerson} \textbf{Dee (born May 18, 1933 in Quincy, Massachusetts) is a former three-sport letterman at Holy Cross College} \\ \textbf{who was one of the first players signed by the Boston Patriots in 1960.} \end{tabular} \\ \bottomrule
\end{tabular}
}
\caption{EEG-to-Text decoding example results on test sentences under five reading tasks. \textbf{Bold} words indicate exact match, \textit{Italic} words indicate semantic resemblance, and \underline{Underline} words indicate error match.}
\label{tab12}
\end{table*}

\clearpage

\begin{table*}[!t]
\vspace{-1cm}
\centering
\resizebox{\textwidth}{!}{
\begin{tabular}{@{}l|llllllllllllllllll@{}}
\toprule
Subjects & YAG & YAK & YMS & YHS & YSL & YRK & YRH & YDR & YIS & YRP & YLS & YTL & YFR & YDG & YAC & YFS & YMD & YSD \\ \midrule
BLEU-1 & 46.23 & 46.67 & 45.65 & 46.12 & 46.50 & 46.34 & 45.90 & 46.13 & 45.90 & 46.45 & 46.12 & 46.56 & 44.75 & 46.78 & 46.28 & 46.51 & 46.89 & 45.65 \\
BLEU-2 & 28.98 & 28.93 & 28.80 & 28.94 & 29.57 & 29.10 & 28.88 & 29.28 & 28.78 & 29.41 & 28.94 & 29.63 & 27.79 & 29.60 & 28.70 & 29.93 & 29.82 & 28.52 \\
BLEU-3 & 18.07 & 17.74 & 17.69 & 17.85 & 18.32 & 17.70 & 17.82 & 18.76 & 17.57 & 18.45 & 17.64 & 18.44 & 16.90 & 18.22 & 17.44 & 18.87 & 18.52 & 17.88 \\
BLEU-4 & 11.27 & 10.85 & 11.09 & 11.04 & 11.22 & 10.70 & 10.89 & 12.10 & 10.64 & 11.82 & 10.67 & 11.44 & 9.88 & 11.34 & 10.50 & 12.09 & 11.48 & 11.18 \\
ROUGE1-R & 35.21 & 35.66 & 35.73 & 34.99 & 36.00 & 35.23 & 35.86 & 35.17 & 34.77 & 35.37 & 35.13 & 35.58 & 34.24 & 34.86 & 35.30 & 35.62 & 35.94 & 35.03 \\
ROUGE1-P & 41.55 & 42.46 & 42.88 & 41.91 & 43.32 & 41.69 & 42.36 & 41.53 & 41.29 & 42.20 & 42.20 & 42.04 & 40.27 & 41.37 & 42.30 & 42.84 & 42.97 & 41.90 \\
ROUGE1-F1 & 38.02 & 38.65 & 38.87 & 38.04 & 39.22 & 38.09 & 38.73 & 37.97 & 37.66 & 38.39 & 38.24 & 38.45 & 36.92 & 37.72 & 38.41 & 38.79 & 39.04 & 38.05 \\ \bottomrule
\end{tabular}}
\caption{Subject-independent Performance of BLEU-N(\%) and ROUGE-1 from Subject YAG to YSD.}
\label{tab13}
\end{table*}

\begin{table*}[h]
\centering
\resizebox{\textwidth}{!}{
\begin{tabular}{@{}l|llllllllllll@{}}
\toprule
Subjects & ZKW & ZPH & ZAB & ZKB & ZMG & ZJN & ZDN & ZJM & ZGW & ZDM & ZKH & ZJS \\ \midrule
BLEU-1 & 37.99 & 38.49 & 38.16 & 38.02 & 37.97 & 38.31 & 37.84 & 38.05 & 38.36 & 38.15 & 38.19 & 37.11 \\
BLEU-2 & 20.83 & 21.07 & 20.83 & 20.89 & 21.14 & 20.74 & 20.81 & 20.73 & 21.58 & 20.92 & 21.00 & 20.34 \\
BLEU-3 & 10.82 & 11.19 & 11.14 & 10.91 & 11.40 & 11.16 & 11.19 & 10.72 & 11.75 & 10.90 & 11.13 & 10.48 \\
BLEU-4 & 5.76 & 6.01 & 6.18 & 5.70 & 6.34 & 6.18 & 6.27 & 5.55 & 6.60 & 5.82 & 6.29 & 5.49 \\
ROUGE1-R & 25.34 & 25.21 & 24.51 & 25.38 & 25.44 & 25.53 & 25.46 & 25.27 & 26.15 & 25.08 & 25.78 & 24.15 \\
ROUGE1-P & 30.44 & 30.43 & 29.39 & 30.74 & 30.55 & 30.48 & 30.31 & 30.27 & 31.14 & 30.10 & 31.02 & 28.84 \\
ROUGE1-F1 & 27.55 & 27.45 & 26.62 & 27.67 & 27.64 & 27.65 & 27.53 & 27.43 & 28.30 & 27.24 & 28.04 & 26.17 \\ \bottomrule
\end{tabular}}
\caption{Subject-independent performance of BLEU-N(\%) and ROUGE-1 from Subject ZKW to ZJS.}
\label{tab14}
\end{table*}

\begin{table*}[h]
\centering
\resizebox{1.0\textwidth}{!}{
\begin{tabular}{@{}c|c|c|cccc|ccc@{}}
\toprule
\multirow{2}{*}{Method} & \multirow{2}{*}{\begin{tabular}[c]{@{}c@{}}Training   \\ Sample\end{tabular}} & \multirow{2}{*}{\begin{tabular}[c]{@{}c@{}}Mask      \\ Stragety \end{tabular}} & \multicolumn{4}{c|}{BLEU-N(\%)} & \multicolumn{3}{c}{ROUGE-1   (\%)} \\
 &  &  & N=1 & N=2 & N=3 & N=4 & P & R & F \\ \midrule
\multirow{2}{*}{E2T-PTR} & 10710 & Random Mask & 40.51 & 24.10 & 14.05 & 8.24 & 35.38 & 29.68 & 32.17 \\
 & 10710 & Force Mask & 42.09 & 25.13 & 14.84 & 8.99 & 35.86 & 30.01 & 35.61 \\ \bottomrule
\end{tabular}}
\caption{Investigating the impact of mask strategy in EEG feature sequences during CET-MAE pre-training.}
\label{tab15}
\end{table*}

\begin{table*}[h]
\centering
\resizebox{1.0\textwidth}{!}{
\begin{tabular}{@{}cc|c|cccc|ccc@{}}
\toprule
\multicolumn{2}{c|}{Model} & \multirow{2}{*}{\begin{tabular}[c]{@{}c@{}}Training   \\ Sample\end{tabular}} & \multicolumn{4}{c|}{BLEU-N(\%)} & \multicolumn{3}{c}{ROUGE-1(\%)} \\
CET-MAE & E2T-PTR &  & N=1 & N=2 & N=3 & N=4 & P & R & F \\ \midrule
{\ding{53}} & Joint   Stream & 10710 & 41.60 & 24.53 & 14.19 & 8.35 & 35.34 & 29.57 & 32.09 \\
{\checkmark} & Joint   Stream & 10710 & 41.61 & 24.57 & 14.34 & 8.52 & 35.74 & 29.79 & 32.37 \\
{\checkmark} & EEG   Stream & 10710 & 42.09 & 25.13 & 14.84 & 8.99 & 35.86 & 30.01 & 32.61 \\ \bottomrule
\end{tabular}}
\caption{We validated the performance impact of multi-stream design on pre-training and downstream tasks. The {\checkmark} indicates the use of a multi-stream design during pre-training, while the {\ding{53}} indicates no use. }
\label{tab16}
\end{table*}

\end{document}